\pdfoutput=1
\documentclass[10pt,twocolumn,letterpaper]{article}

\usepackage{cvpr}              % To produce the CAMERA-READY version

\usepackage{graphicx}

\usepackage{amsmath}
\usepackage{amssymb}
\usepackage{booktabs}
\usepackage{subfloat}
\usepackage{multirow}
\usepackage[colorlinks,linkcolor=red]{hyperref}
\usepackage[numbers, sort&compress]{natbib}
\usepackage{threeparttable}
\usepackage{algorithm}
\usepackage{algpseudocode}
\usepackage{amsmath}
\usepackage{graphics}
\usepackage{epsfig}
\usepackage{amsthm}
\usepackage[capitalize]{cleveref}
\crefname{section}{Sec.}{Secs.}
\Crefname{section}{Section}{Sections}
\Crefname{table}{Table}{Tables}
\crefname{table}{Tab.}{Tabs.}

\def\cvprPaperID{10058} % *** Enter the CVPR Paper ID here
\def\confName{CVPR}
\def\confYear{2022}

\begin{document}

%%%%%%%%% TITLE - PLEASE UPDATE
\title{How to Explain Neural Networks: an Approximation Perspective}

% \author{Jingxiao Liao, Donghang Cheng, Jinwei Sun, Shiping Zhang$^*$\\
% Harbin Institute of Technology, Harbin, China\\
% {\tt\small jingxiaoliao@hit.edu.cn}
% % For a paper whose authors are all at the same institution,
% % omit the following lines up until the closing ``}''.
% % Additional authors and addresses can be added with ``\and'',
% % just like the second author.
% % To save space, use either the email address or home page, not both
% \and
% Fenglei Fan$^*$\\
% Institution2\\
% {\tt\small secondauthor@i2.org}
% }

\author{Hangcheng Dong$^{1}$, Bingguo Liu$^{1}$, Fupeng Wei$^{2}$, Fengdong Chen$^{1}$, Dong Ye$^{1}$, and Guodong Liu$^{1*}$\\
        $^{1}$Harbin Institute of Technology, Harbin, China\\
        $^{2}$North China University of Water Resources and Electric Power, Zhengzhou, China\\
        {\tt\small \{hunsen\_d, lgd\}@hit.edu.cn}
        }

\maketitle

%%%%%%%%% ABSTRACT
\begin{abstract}
The lack of interpretability has hindered the large-scale adoption of AI technologies. However, the fundamental idea of interpretability, as well as how to put it into practice, remains unclear. We provide notions of interpretability based on approximation theory in this study. We first implement this approximation interpretation on a specific model (fully connected neural network) and then propose to use MLP as a universal interpreter to explain arbitrary black-box models. Extensive experiments demonstrate the effectiveness of our approach.

\end{abstract}

%%%%%%%%% BODY TEXT
\section{Introduction}
The interpretability of deep learning plays a key role in many mission-critical applications~\cite{lipton_mythos_2018}. In the past years, interpretability has been an active research field of deep learning. Currently, studies on interpretability~\cite{Fan2021,zhang_survey_2021,BarredoArrieta2020,Adadi2018,Samek2021,Vilone2020} can be divided into two categories: post-hoc explanation and ad-hoc explainable model prototyping~\cite{Rudin2019,interpretable-cnn,self-explainingNN,prototypesNN}. Post-hoc explanation methods attempt to offer an explanation for a model when the model is trained, while the ad-hoc explanation methods prototype a more interpretable model from the beginning. Despite that these studies have accomplished great successes, we argue that the fundamental question on interpretability remain unanswered, i.e., what kind of interpretability do we want? what kind of interpretability models can succeed? and how to realize them?

% In the post-hoc explanation methods, feature visualization~\cite{ancona2018towards} and building a proxy for target black-box models~\cite{distill} are the primary approaches. 

% In terms of developing explainable models, there are less constraints. 

\begin{figure}[t]
\centering
\includegraphics[scale=0.3]{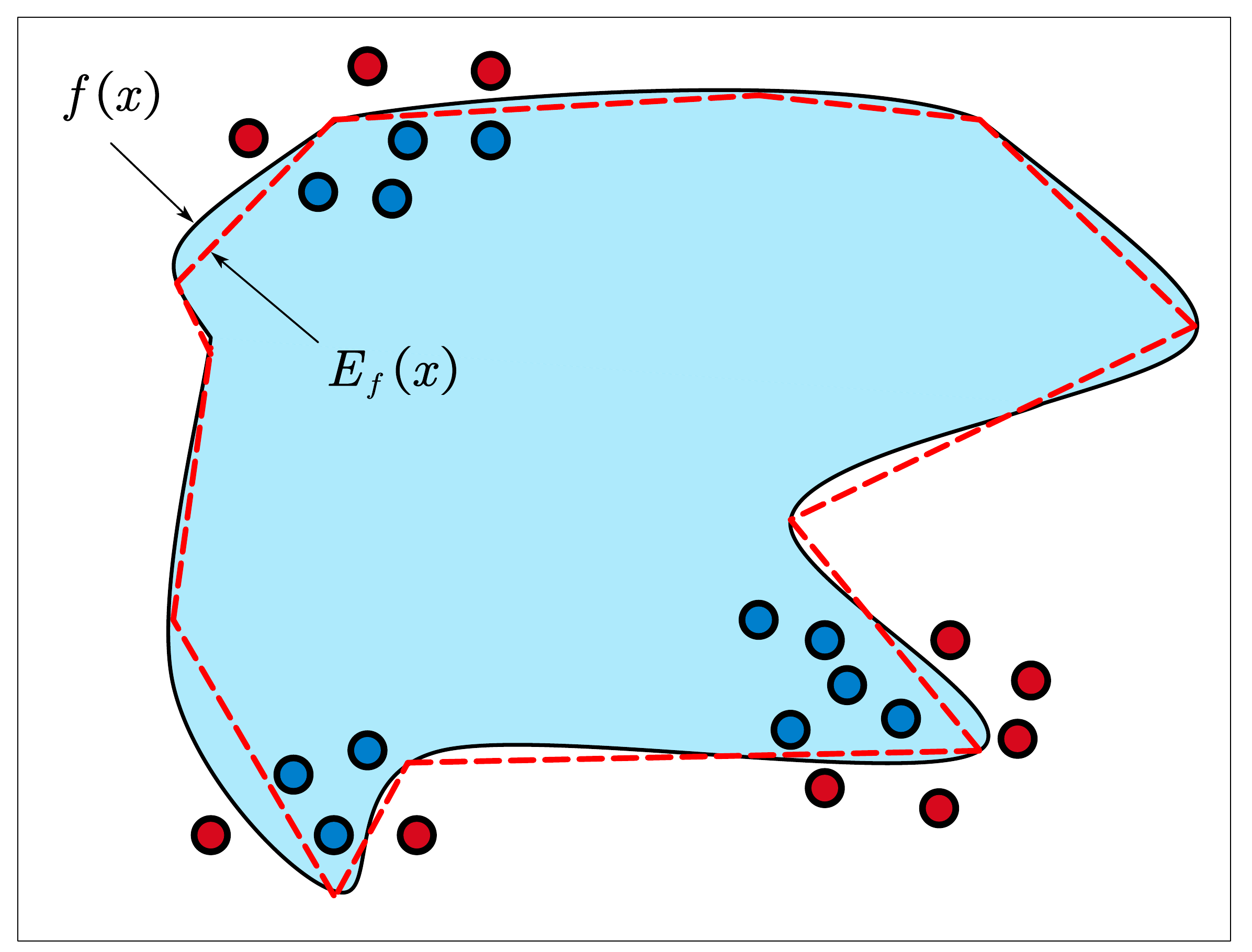} 
\caption{Explain black box models by the approximation mechanism. $f(x)$ is a black-box model, and $E_f(x)$ is a white-box model.}
\label{fig1}
\end{figure}

To answer the above questions, we must figure out the desirable characters we want an explanation to have. We think that an explanation shall fulfill \textbf{completeness}~\cite{saliency6,Srinivas:272017}, i.e., the explanation should be faithful to the origin model not locally but in the entire domain. Unfortunately, the majority of existing approaches do not respect the completeness when generating an explanation well. For example, gradient-based saliency maps are usually incomplete since they are essentially based on changes of model outputs without taking the model bias into account. Other than this, we think that the major difficulty of gaining completeness is because too many parameters are interacted complicatedly in a black-box model, rendering it hard to analyze the influence of each parameter on the output of the model. 

Our motive is to develop an interpretabibility method to accomplish completeness. Recently, some researches~\cite{Avner1996ExtractionOC,67354,hanin2019deep,Zhang2020Empirical,chu_exact_2018,LEI2020361,Srinivas:272017} have made huge strides in interpretability by reducing a fully-connected ReLU network into a piecewise linear function over polytopes. Given such a ReLU network, for any input, we can identify the function over its pertaining polytope, thereby favorably providing a complete explanation. Since a fully-connected ReLU network has a complete explanation in the light of piecewise linear functions, our curiosity is can we use a fully-connected ReLU network to explain other black-box neural networks? Specifically, can we distill other black-box neural networks into a fully-connected ReLU network to gain interpretability?    

Such an idea is desirable due to the fact that, to the best of our knowledge, there is still a vacuum in global completeness of interpretation methods. We consider interpretability to be an approximation problem. Simultaneously, the approximation procedure is necessary to ensure completeness (refer Fig.\ref{fig1}). Therefore, we present the concept of CLIME (complete local interpretable model-agnostic explanations), which is the global formulation of LIME~\cite{Ribeiro2016}. In this research, we first focused at how to understand any fully-connected neural network (FCNN), and then we introduced an idea of distillation-based CLIME to explain arbitrary black-box models. Our study made the following significant contributions:

(1) We presented the definition and properties of an explanation, which is called CLIME. Furthermore, we show that in the classification problem, CLIME is equivalent to the decision boundary of black-box models.

(2) We explained FCNN by finding the piecewise linear approximation, which is an extended version of the technique in ~\cite{chu_exact_2018}. It is worth to point out that a similar method named LANN (linear approximation neural network) has been proposed in~\cite{huxia}. The main difference is the approximation method of activation functions. Moreover, our goal is to compute approximation decision boundaries,  whereas theirs is to measure the model complexity. 

(3) We proposed DecisionNet (DNet), a wide network for binary classification, as a white-box representation of PLNN. This transformation allows the parameters of PLNN to be decoupled, which meets our definition of explainable models.

(4) We proposed to explain other black-box models using PLNNs (FCNNs), i.e., a preliminary implementation of CLIME based on knowledge distillation.

\section{Related Works \label{sec:related works}}

\textbf{Feature Visualization} is a straightforward method for interpretability~\cite{ancona2018towards}, whose idea is to generate saliency maps depicting the concerned parameters. Observing activation maps~\cite{2015Understanding} is one of the most intuitive notions. However, this strategy has shown limited results. Exploration of the link between activation maps and the input space, also known as activation maximization~\cite{erhan2009visualizing} and deconvolution~\cite{Dconv}, is another consideration. Both of them have demonstrated that as the number of layers in a well-trained network grows, the semantic information on which neurons are focused becomes more richer. A number of gradient-based approaches~\cite{saliency1,saliency2,saliency3,saliency4,saliency5,saliency6,saliency7,saliency8} have been presented to quantify the relevance of input characteristics, and perturbation-based~\cite{pt1,pt2,pt3} methods have also attracted a lot of attention. These technologies have shown that neural networks can learn semantic information, but they are susceptible and sensitive to small changes in the input space~\cite{35}. CAM (Class Activation Mapping)~\cite{36cam} and its derivatives~\cite{gradcam,gradcamplus,scam} are other well-known visualization approaches. Heatmaps are created using linear combinations of activation maps from convolution layers in CAM-like techniques. However, as previously discussed, feature visualization approaches are not complete.

\textbf{Proxy Method} tries to train simple models based on the black-box model. Hinton et al. demonstrated how knowledge distillation might increase generalization performance by leveraging the output of a big neural network, known as soft targets, to train another model (such as a tiny neural network or decision trees). Due to a shortage of efficient explainable student models, model distillation is frequently employed to improve the performance of them~\cite{Frosst2017,distill,KDTree,gou2021knowledge}. LIME~\cite{Ribeiro2016} based on local approximation are presented to solve the problems of global proxy. LIME retrains a linear classifier to act as a local proxy for the origin model by sampling in the neighborhood of the sample to be explained. On the other hand, LIME can only understand models locally and is difficult to execute on data with high dimensions. We propose to use PLNN as an explainable model that enables the implementation of CLIME, which can combine the benefits of global and local proxy methods.

\textbf{ReLU Network} is a piecewise linear function, which divide the input space into several linear regions~\cite{67354,hanin2019deep,chu_exact_2018}. The upper bound on the number of the linear regions has been analyzed in ~\cite{67354,hanin2019deep,LEI2020361}. The influence of BN and dropout strategies on the linear interval of ReLU networks was also observed in the literature~\cite{Zhang2020Empirical}. Hu~\cite{huxia} el at. proposed linear approximation neural network by using piecewise linear functions instead of curve activation functions. Inspired by these studies, we believe that the piecewise linear function satisfies both completeness and explicitness, making it a quiet perfect explainable model.

%-------------------------------------------------------------------------
\section{Definitions of Interpretability \label{sec:properties}}

In this section, we discuss the following questions: 1) What is the explainable model? 2) What is interpretability? 3) How to explain black-box models?

We contend that there are two levels of interpretability: the first is the explanation of external appearances, and the second is the comprehension of inner mechanisms. We believe that because a ReLU network is a piecewise linear function, it is hard to describe the intrinsic mechanism of a neural network, as the intrinsic mechanism most likely does not exist. As a result, the explainable model is defined as follows:

%\newtheorem{definition}{Definition (sadsa)}
%\begin{definition}
%If a model $E_f(\mathbf{x};\boldsymbol{\theta})$ is explainable, implies that we are able to know the %\textbf{global} impact of each parameter $\boldsymbol{\theta}_i$ on the output, which termed %explicitness.
%\end{definition}

\textbf{Definition 1 (Explainable Model)}.  \emph{If a model $E_f(\mathbf{x};\boldsymbol{\theta})$ is explainable, implies that we are able to know the \textbf{global} impact of each parameter $\boldsymbol{\theta}_i$ on the output, which termed explicitness.}  

We would like to to underline that the derivative itself is insufficient to indicate the global influence of a parameter. Kernel SVM~\cite{PRML}, ensemble Learning~\cite{zhou2021ensemble}, and deep learning~\cite{goodfellow2016deep} are black-box, while (piecewise) linear models are white-box, according to Definition 1. Based on the explainable model, we further define interpretability according to approximation theory.

\textbf{Definition 2 (Model Interpretation)}.  \emph{For a given error $\delta>0$ and model $f(\cdot)$, if there exist an explainable model $E_f(\cdot)$, we say $E_f(\cdot)$ is an interpretation of $f(\cdot)$, implying that it satisfies the following conditions:  
\begin{equation}
    \forall \mathbf{x} \in \chi\subseteq \mathbb{R}^d, \vert f(\mathbf{x})-E_f(\mathbf{x})\vert<\delta,
    \label{eq1}
\end{equation}
where $\chi$ is the input space, $\mathbf{x}$ and $d$ are an instance and the dimension of input features respectively.}

We define interpretation as an approximation process employing a specific white-box model, which implies that the precision that an interpretation may attain is heavily determined by the white-box model utilized. The white-box model that fits Definition 2 is referred to as the complete local interpretable model-agnostic explanation (\textbf{CLIME}). However, because achieving a model-agnostic explanation is challenging, we will first discuss implementing in a specific model, namely the FCNN. Following that, we will explore deployment in any model. In this work, they are all referred to as CLIME without distinction.

\section{Interpretability of Classification Models}

In this section, we introduce the relationship between CLIME and the dicision boundary corresponding to the classification model. The decision boundary of a classifier is defined as follows.

\textbf{Definition 3 (Decision Boundary)}.\emph{ Given a classifier $F(\mathbf{x})=argmax_i(f_1(\mathbf{x}),f_2(\mathbf{x}),...,f_i(\mathbf{x}),...,f_d(\mathbf{x}))$ where $f:\mathbb{R}^{d} \rightarrow \mathbb{R}^{c}$ and $f_i(\mathbf{x})$ is the $i$-th value of $f(\mathbf{x})$. Denote by $C=\{C_i \vert i=1, 2, ..., n\}$ the labels. The decision boundary between class $C_i$ and $C_j(i\neq j)$ is: 
\begin{equation}
    DB_{ij}=\{\mathbf{x}\vert\forall \delta >0, \exists \mathbf{x}_i, \mathbf{x}_j\in U(\mathbf{x}, \delta), f(\mathbf{x}_i)=f(\mathbf{x}_j)\},
    \label{eq2}
\end{equation}
where $U(\mathbf{x},\delta)$ means a open region satisfying $U(\mathbf{x},\delta)=\{z \vert dist(\mathbf{x},z)<\delta\}$, $dist(\cdot,\cdot)$ is the distance metric.}

Next, we extend Definition 2 to the classification problem.

\textbf{Definition 4 (Classifier Interpretation)}.\emph{ For a given classier $F(\cdot)$, if there exist an explainable model $E_F(\cdot)$, we say $E_F(\cdot)$ is a classifier interpretation of $F(\cdot)$, implying that it satisfies the following conditions:
\begin{equation}
    \forall x\in \chi\subseteq\mathbb{R}^d, F(x)=E_F(x).
    \label{eq3}
\end{equation}
}
\textbf{Proposition 1}. \emph{When the error $\delta$ defined in Definition 2 is small enough, we have that Definition 2 is a sufficient condition for Definition 4.}

\begin{proof}
 Let the classifier be $F(\mathbf{x})$, and the $i$-th output of logit $f_i(\mathbf{x})$ be the maximum, namely $f_i-f_j>0, (\forall j \in {k \vert 1, 2, ..., n, k\neq i})$. Denote by $E_F(\mathbf{x})$ the explanation. Notice the completeness by Definition 1 of $E_f(\cdot)$, we have:
\begin{equation}
    \begin{aligned}
    &\underset{\delta\rightarrow0}{lim}(E_{f_i}-E_{f_j})= \\
    &\underset{\delta\rightarrow0}{lim}[(E_{f_i}-f_i)+(f_i-f_j)+(f_j-E_{f_j})]=f_i-f_j>0.
    \end{aligned}
    \label{eq5}
\end{equation}
Thus, $\exists \delta>0,\mathbf{x} \in U(\mathbf{x},\delta) $, we have $E_{f_i}(\mathbf{x})-E_{f_j}(\mathbf{x})>0, (\forall j \in {k \vert 1, 2, ..., n, k\neq i})$, namely, $E_F(x)=F(x)$, which concludes the proof.
\end{proof}

In consequence we still call the classifier interpretation as CLIME. At last, we would like to show the relationship between CLIME and the corresponding decision boundary in the scenario of a classification problem.

\textbf{Theorem 1}.\emph{ For a classifier, the classifier interpretation (i.e. CLIME) is equivalent to the corresponding decision boundary.}

\begin{proof}
 In a classification problem, the necessity is that, for a trained classifier  $F:\mathbb{R}^{d} \rightarrow \mathbb{R}$, given the decision boundary, the input space $\chi\subseteq \mathbb{R}^d$ can be divided into several maximum decision regions $R_i=\{\mathbf{x}\vert F(\mathbf{x})=i (i \in {1,2,...,n})\}$, then CLIME $E_F(\cdot)$ can be described as $\{R_i\}_1^n$.
 
The sufficiency is that, $\forall \mathbf{x} \in \chi$, by the Proposition 1, CLIME $E_F(\cdot)$ is a classifier interpretation, that is $F(\mathbf{x})=E_F(\mathbf{x})$, then the decision boundary set is $\{DB_{ij}\vert i,j=1,2,...,n,i\neq j\}$, where $DB_{ij}=\underset{\mathbf{x} \in R_i\cap R_j}{\bigcup}(\mathbf{x})$, which concludes the proof.
 
\end{proof}

%%%%%%%%%%%%%%%%%%%%%%%%%%%%%%%%%%%%%section5%%%%%%%%%%%%%%%%%%%%%%%%%%%%%%%%%
\section{Interpret FCNNs}

In this section, we show how to obtain CLIME of an FCNN with common activation functions. First, the FCNN with piecewise linear activation function (PLNN) and its properties are discussed. Second, we demonstrate how to calculate CLIME for an FCNN with a nonlinear activation function. The computation of the approximation decision boundary corresponding to an FCNN is then shown. Finally, we look into how to transform an FCNN into a white-box model.

\subsection{Interpretability of PLNNs}

\begin{figure}[htpb]
\centering
\includegraphics[scale=0.25]{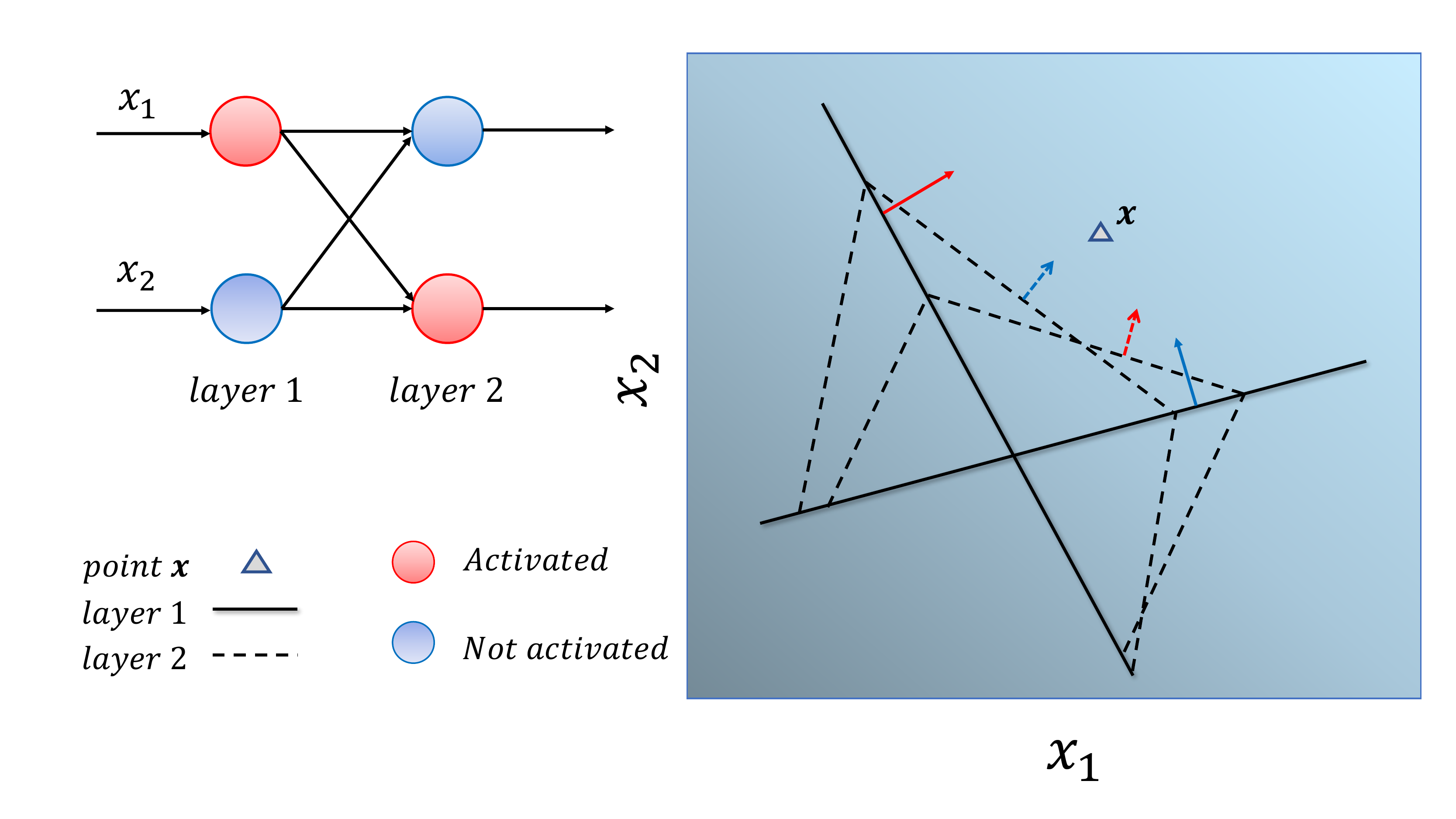} 
\caption{A hyperplane in data space correlates to the neuron's activation state. The first layer of neurons will split the whole space, the second layer will divide all of the first layer's subspaces except the preceding layer's dead zone (all neurons are not active), and so on.}
\label{fig2}
\end{figure}

A PLNN can divide the input space into several linear regions, since it is a composite of continuous piecewise linear functions. Samples with the same activation state will be in the same linear interval. Activation state means that on each neuron, the sample is activated by the same linear region of the activation function, as illustrated in Fig.\ref{fig2}.

According to the linear regions of PLNNs, we can naturally obtain the CLIME. On each linear interval $\Omega _i$, CLIME can be expressed as the following formula~\cite{chu_exact_2018}:
\begin{equation}
    \forall \mathbf{x} \in \Omega _i, E_{f}(\mathbf{x})=(\partial f/\partial \mathbf{x})^T\mathbf{x}+ (\partial f/\partial b)^T b.
    \label{eq6}
\end{equation}

However, the time complexity of scanning all of the activation state is $O(k^n)$, where $k$ is the number of segments of the activation function, and $n$ is the number of all the hidden neurons. Thus, we can use the OpenBox method~\cite{chu_exact_2018}, which reduces the complexity by querying samples. 

In summary, PLNN can be equivalently approximated by a piecewise linear model. However, since it relies on sample queries, we term it the \textbf{lazy explainable model}.

\subsection{Interpretability of FCNNs}

We extend the interpretability of PLNNs to FCNNs, motivated by their interpretability. We propose piecewise linear activations $p_n(x)$ as an alternative to the commonly utilized activation functions $\sigma(x)$, which are generally Lipschitz continuous and have infinite asymptotes. We require $p_n(x)$ to meet the following condition to satisfy Definition 2:
 
\begin{equation}
    \underset{n\rightarrow \infty}{lim}(\sigma(x)-p_n(x))=0,
    \label{eq7}
\end{equation}
where $n$ is the number of segments.

\begin{figure}[!htpb]
\centering
\subfloat[sigmoid]{
\label{Fig.sub.1}
\includegraphics[scale = 0.22]{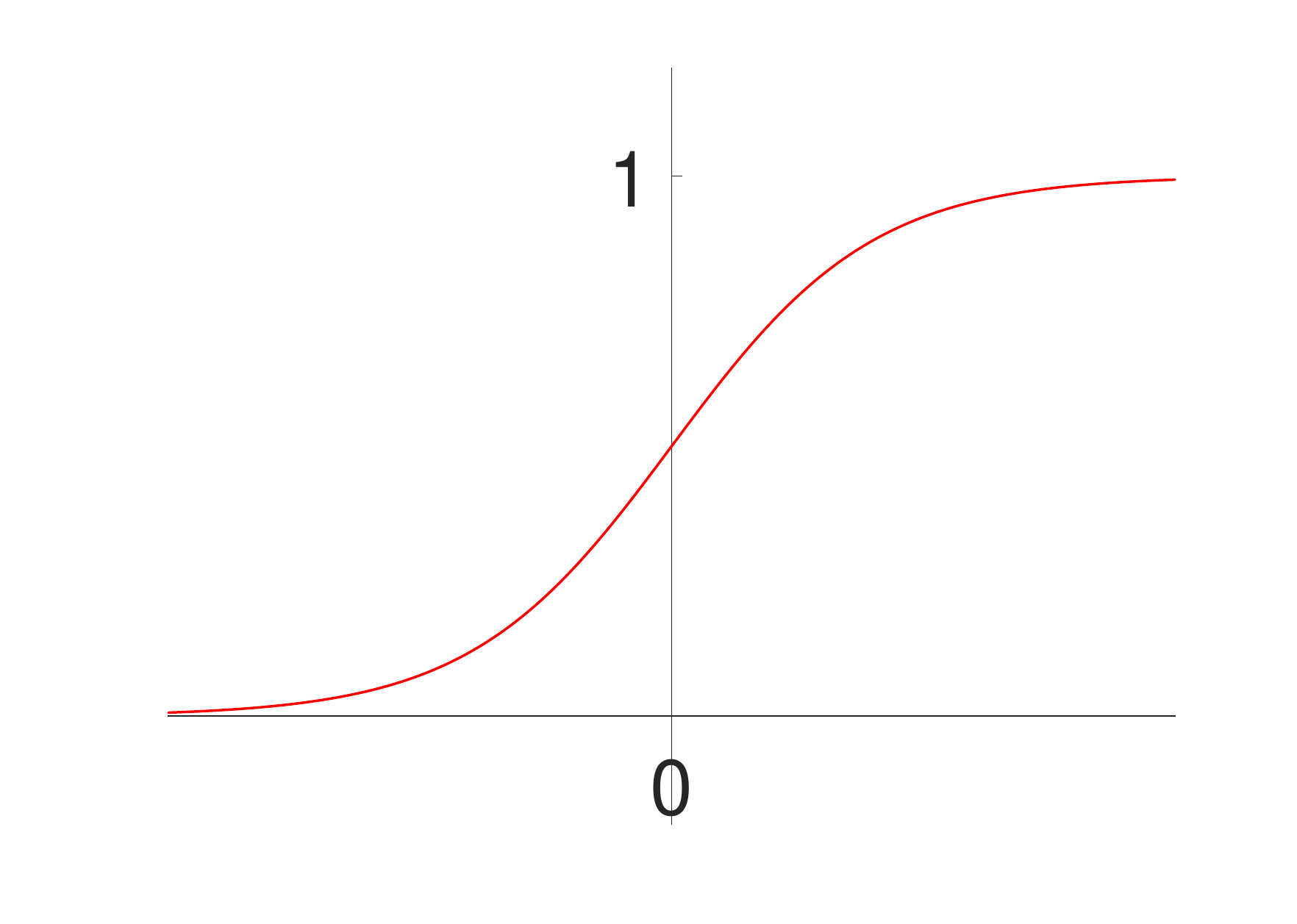}}
\subfloat[$p_5$-sigmoid]{
\label{Fig.sub.2}
\includegraphics[scale = 0.22]{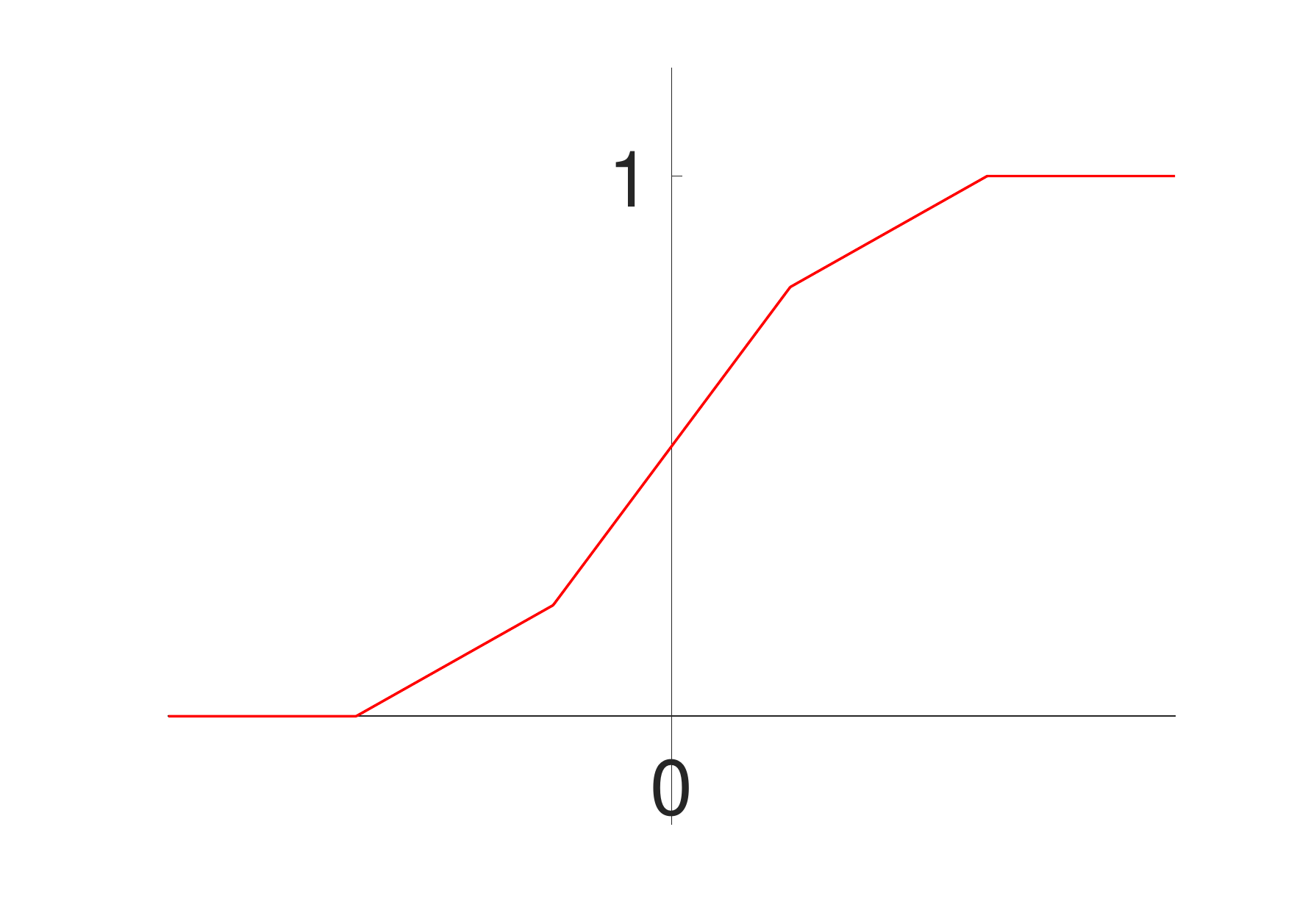}}
\caption{The piecewise linear approximation with $5$ segments of the sigmoid function calculated by Algorithm 1.}
\label{fig3}
\end{figure}

Algorithm 1 summarizes the construction process of $p_n(x)$. The linear approximation of sigmoid function is described in Fig.\ref{fig3}. By the linear approximation, the CLIME of FCNNs can be obtained as the same as PLNNs. A remaining question is whether the scheme meets Definition 2. 

For simplicity, consider an FCNN with two layers, $f:\mathbb{R}^d \rightarrow \mathbb{R}^c$. The layers are noted as $f_1(\cdot)$ and $f_2(\cdot)$ respectively, and the approximation layers (activations are generated from Algorithm 1) are noted as $g_1(\cdot)$ and $g_2(\cdot)$ respectively. For an instance $\mathbf{x}\in \mathbb{R}^d$, we have:

\begin{equation}
\begin{aligned}
&\Vert f_1(f_2(\mathbf{x}))-g_1(g_2(\mathbf{x})) \Vert =  \\
&\Vert f_1(f_2(\mathbf{x}))-g_1(f_2(\mathbf{x}))+g_1(f_2(\mathbf{x}))-g_1(g_2(\mathbf{x})) \Vert \\
 &\le \Vert f_1(f_2(\mathbf{x}))-g_1(f_2(\mathbf{x}))\Vert+\Vert g_1(f_2(\mathbf{x}))-g_1(g_2(\mathbf{x})) \Vert
\end{aligned}
\label{eq8}
\end{equation}

Considering the Lipschitz continuity of $f_1, f_2, g_1, g_2$, we have:
\begin{equation}
    \begin{aligned}
        \Vert f_1(f_2(\mathbf{x}))-g_1(g_2(\mathbf{x})) \Vert \le C \cdot \delta(n),
%       \Vert g_1(f_2(\mathbf{x}))-g_1(g_2(\mathbf{x})) \Vert &\le L\cdot w\cdot \Vert f_2(\mathbf{x})-g_2(\mathbf{x})\Vert\\
%   & =L\cdot w\cdot \delta(n)
    \end{aligned}
    \label{eq9}
\end{equation}
where $C$ is a constant and $\underset{n\rightarrow \infty}{lim}\delta(n)=0$. It concludes that the approximation scheme meets Definition 2.

%\begin{equation}
%    \Vert f_1(f_2(\mathbf{x}))-g_1(g_2(\mathbf{x})) \Vert\le (C(L,w)+1)\cdot\delta(n)
%    \label{eq10}
%\end{equation}

In summary, we can transform an FCNN into a PLNN by linearizing the activations, which means that we can obtain the interpretation of an FCNN.

\begin{algorithm}[t]
  \caption{Linear approximate activation function} % 名称
  \label{alg::conjugateGradient}
  \begin{algorithmic}[1]
    \Require
     $\sigma(x):=$ the target activation function, $n:=$ the number of pieces, ($n \geq n_0$), $n_0$:=the number of 
asymptotic lines of $\sigma(x)$.
    \Ensure
      $p_n(x)$:=a piece-wise linear function with $n$ pieces
    \State Initialization: caulate the hard-$\sigma(x)$ which is consist of the asymptotic lines of $\sigma(x)$, $p_0(x)=$hard-$\sigma(x)$
    \State \textbf{for} $i$ in range(1,$n-n_0+1$) \textbf{do}
    \State \quad compute the point by $x \leftarrow argmax_x(\sigma(x)-p_n(x))$
    \State \quad \textbf{if} $\# x> 1$ \textbf{then}
    \State \qquad $x \leftarrow min(x)$
    \State \quad \textbf{end if}
    \State \quad compute tangent line $l_i$ at point $x$  
    \State \quad $p_{n} \leftarrow$ add new line $l_i$ to $p_n$
    \State \textbf{end for}
    \State \textbf{return} $p_n$.
  \end{algorithmic}
\end{algorithm}

%%%

\subsection{Decision boundaries of FCNNs}

In addition to the interpretability of FCNNs, we further desire to understand its CLIME in classification problems, i.e., the decision boundaries. However, it remains an open problem to calculate the decision boundary of black-box models. Existing methods often rely on adversarial samples to estimate the decision boundary~\cite{db1,db2,db3,db4,db5}. In~\cite{chu_exact_2018}, a method for computing the decision boundary of PLNNs has been given. Moreover, we can implement the technology in~\cite{chu_exact_2018} to obtain the decision boundary of FCNNs based on the interpretability of FCNNs. We would like to emphasize that FCNN is still a lazy explainable model, and its calculation of decision bounds is sample-dependent. 

%According to the analysis of decision boundaries, there will be a lot of redundancy in the parameters 5of FCNNs (PLNNs), so we wonder if there exists a more efficient and direct way to express FCNNs %(PLNNs)? We will talk about it in next subsection.

%\begin{algorithm}[t]
%  \caption{Calculating the Decision Boundary} % 名称%
%  \label{alg::conjugateGradient}
%  \begin{algorithmic}[1]
%    \Require
%     $f(x)$:=the target FCNN, $\sigma(x)$:=activation function used in $f(x)$, $\delta$:=given precision
%    \Ensure
%     $DB$:=the decision boundary of $f(x)$
%    \State Initialization:$n=n_0$, $\epsilon=\epsilon_0$, $DB=\emptyset$
%    \State \textbf{while} $\epsilon>\delta$ \textbf{do}:
%    \State \quad$n \leftarrow n+1$
%    \State \quad$p_n(x),\epsilon \leftarrow$ Algorithm 1($\sigma(x),n$)
%    \State \textbf{end while} 
%    \State compute the activation state set $AS \leftarrow \{AS_i:x, f(x)\}$ 
%    \State \textbf{for} each $AS_i$ in $AS$ \textbf{do}
%    \State \quad\textbf{if} the label $f(x)$ in $AS_i$ are not the same \textbf{do}
%    \State \qquad compute the CLIME by Eq.(9)
%    \State \qquad compute the $DB_i$ by Definition 4
%    \State \quad\textbf{end if}
%    \State \quad$DB \leftarrow DB \bigcup DB_i$
%    \State \textbf{end for}
%    \State \textbf{return} $DB$.
%  \end{algorithmic}
%\end{algorithm}

\subsection{DecisionNet : Make Black Box White}

\begin{figure}[!htpb]
\centering
\includegraphics[scale=0.25]{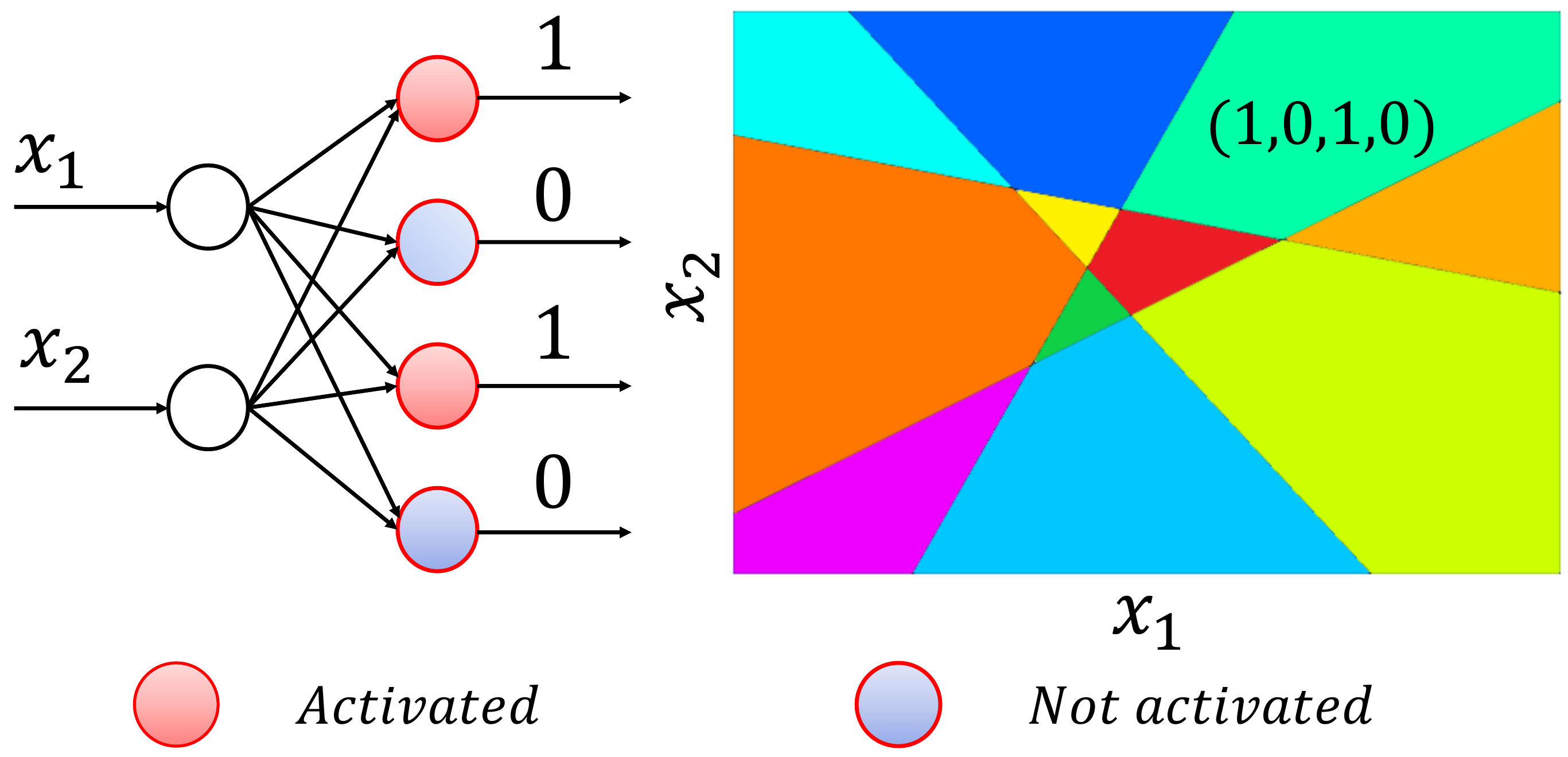} 
\caption{DNet is constructed with parameters of a black-box model's decision boundaries,  which means that each parameter has a clear geometric meaning. For convenience, we display the DNet with ReLU activation, where 0 and 1 means the input is located on the negative and positive axes respectively.}
\label{fig4}
\end{figure}

We propose the DecisionNet (DNet), which is a totally white-box representation of corresponding PLNN that fits Definition 1. The architecture of the DNet, which is a network with only one hidden layer, is shown in Fig.\ref{fig4}. The parameters of DNet are copied from all of the decision boundary hyperplanes. The output of DNet is a vector, with the sign of each component reflecting the input's relative location to the related hyperplane. If the output of DNet is exactly the same in the input space, then the label of such samples must be the same. According to this characteristic, we first use the training set to mark all states of outputs. By comparing the state of an output, we can obtain the label. When a new state emerges that is not covered by the training set, we can rely on the original network for recalibration. 

In addition,  DNet is more suitable as a white-box representation of a binary classification model. DNet decouples the parameters by turning the deep network into a width network. In some simple cases, DNet can save the number of neurons, while the saving of parameters depends more on the dimensionality of the input data.

%%%%%%%%%%%%%%%%%%%%%%%%%%%%%%%%%%%%%图1%%%%%%%%%%%%%%%%%%%%%%%%%%%%%%%%%
%\begin{figure*}[htbp]
    %\centering
    %\includegraphics[]{figure/FCQAEstructure.pdf}
    %\caption{An overview of fully-connected Quadratic AutoEncoder. The core point of it is the replacement of a %single neuron in the hidden layer with a quadratic neuron. The yellow neurons(hidden layers) on the left %side of the figure are all quadratic neurons.}
%\label{fig:QFAEstructure}
%\end{figure*}

\section{Distillation-based CLIME}

Currently, there is a huge performance gap between white-box and complicated models, resulting in proxy method failure. We propose to use PLNNs (FCNNs) to explain other black-box models, considering remarkable abilities of neural networks. For details, we propose to train PLNN (FCNN) students directly using the knowledge distillation technology in ~\cite{distill}. Then, the CLIME of any black-box model can be calculated based on the interpretability of PLNNs (FCNNs).

We would like to point out that distillation-based CLIME is only a preliminary scheme and does not exactly meet the demand. One reason is that the performance of FCNN still needs to be improved on large-scale data, and the other is that it is hard to obtain a precise CLIME by distillation methods.

\section{Experiment}

In this section, we verify the effectiveness of the proposed strategy through systematic experiments. We would like to address the following questions: (1) Is the piecewise linear approximation explanation of FCNN consistent? (2) How is the performance of DNet? (3) How is the effectiveness of distillation-based CLIME? We used 2 toy datasets~\cite{scikit-learn} and 2 commonly-used datasets~\cite{mnist,fashionmnist}. The experiments are performed on a PC with an Intel Xeon Gold 6240 CPU and 4 Nvidia Geforce RTX 2080 Ti GPU.

\label{sec:Experiment}
\subsection{Interpretability of FCNNs}

\begin{figure}[!htpb]
\centering
\subfloat[sigmoid]{
\label{Fig.sub.1}
\includegraphics[scale = 0.24]{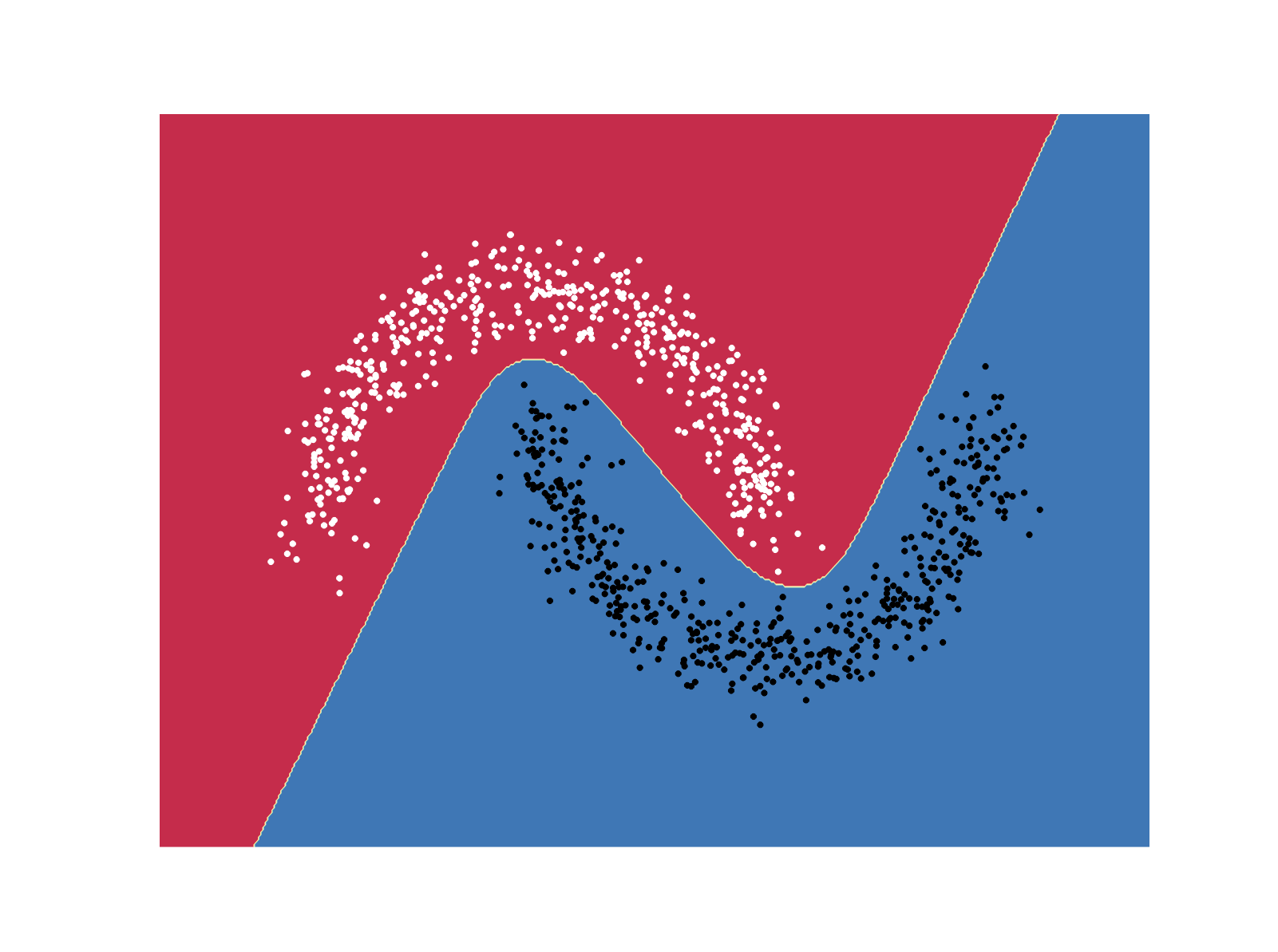}}
\subfloat[$p_3$-sigmoid]{
\label{Fig.sub.2}
\includegraphics[scale = 0.24]{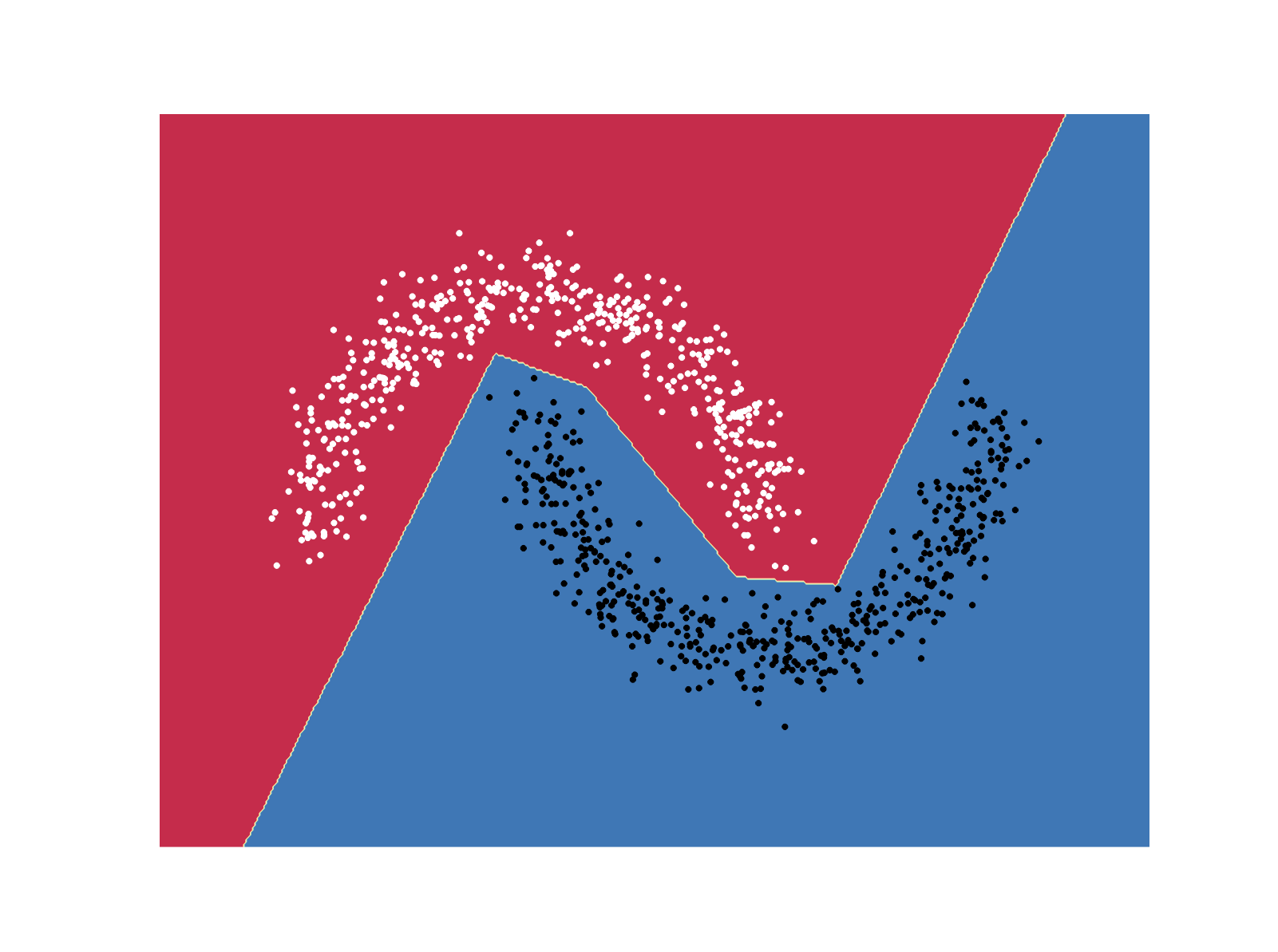}}
\quad

\subfloat[tanh]{
\label{Fig.sub.3}
\includegraphics[scale = 0.24]{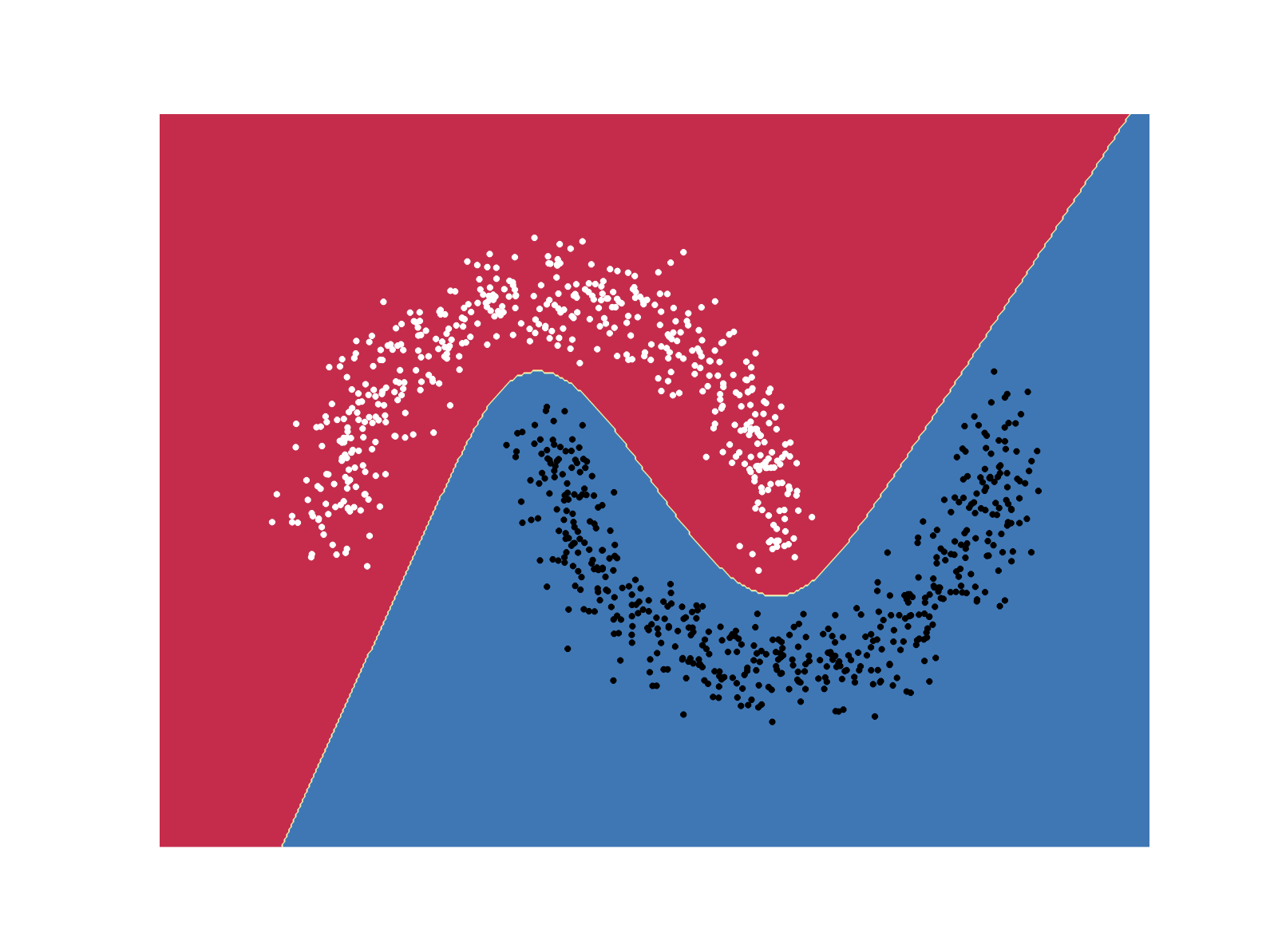}}
\subfloat[$p_3$-tanh]{
\label{Fig.sub.4}
\includegraphics[scale = 0.24]{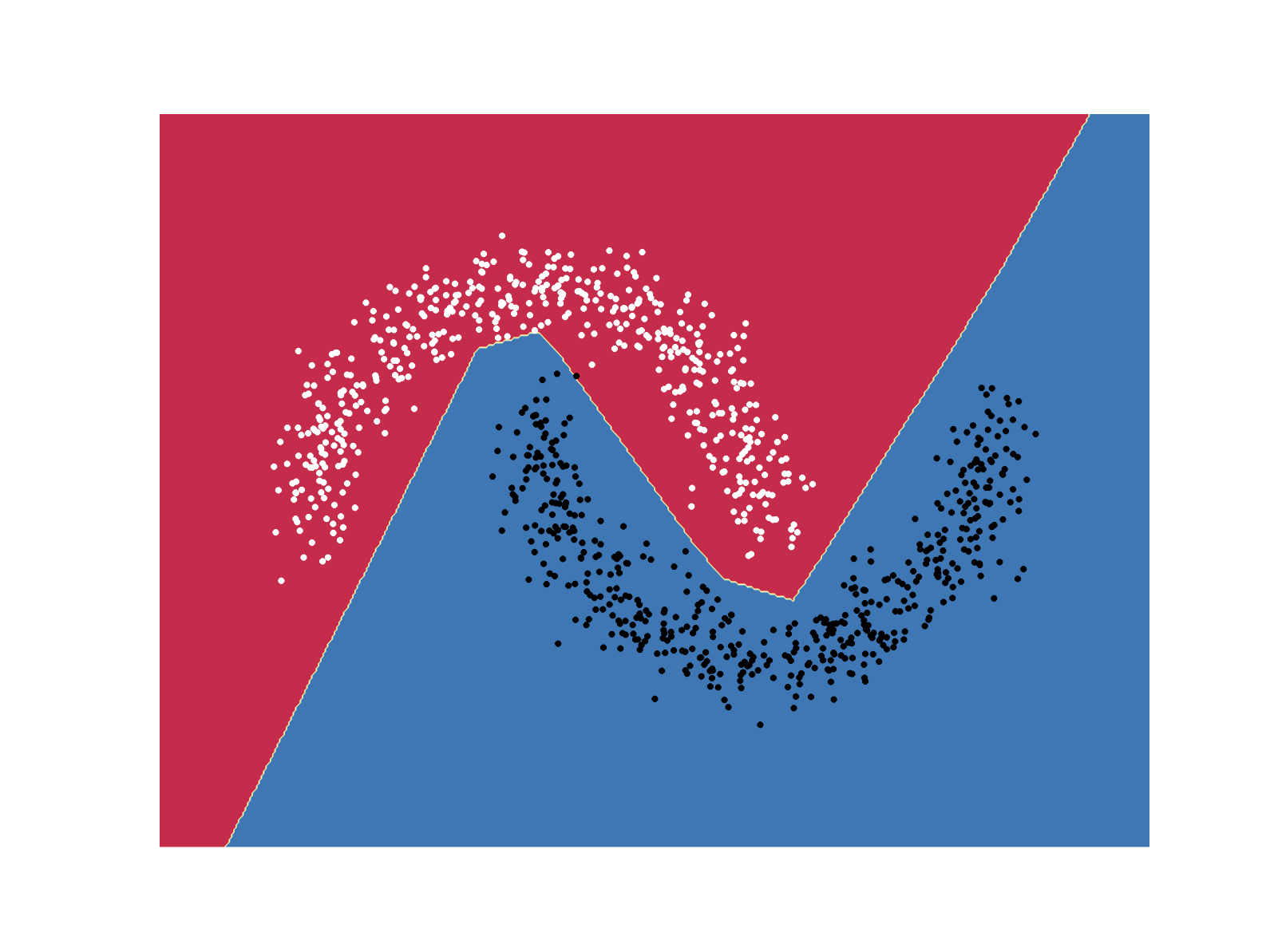}}
\caption{(a) and (c) are the decision boundary of the FCNN with sigmoid and tanh functions, respectively, (b) and (d) show the decision boundary of the FCNN with $p_3$ linear approximate function}
\label{fig5}
\end{figure}

Fig.\ref{fig5} illustrates the results of decision boundary of an FCNN on a toy dataset~\cite{scikit-learn}. We trained an FCNN (2-8-5-2) with sigmoid and tanh activation functions respectively, and used Algorithm 1 to compute their linear approximation by taking the number of pieces $n=3$ (noted as $p_3$-sigmoid/tanh). The results in the figure show that our method is highly efficient.

We trained FCNNs (784-256-64-10) on FashionMNIST dataset using sigmoid, tanh and elu as activation functions respectively, and calculated their approximation PLNNs according to the subsection 5.2. As shown in Table \ref{tab1}, the accuracy of the approximation PLNNs tend to increase as the number of segments $n$ increases. Sometimes we can also find that the accuracy of the approximation PLNN exceeds that of the original model.

\begin{table}[!htbp]
  \centering
  \caption{Accuracy (\%) of FCNNs and approximation models on FashionMNIST dataset.}
    \begin{tabular}{lllll}
    \toprule
        \textbf{activations}  & \textbf{baseline} & \textbf{$n=2$} & \textbf{$n=3$} & \textbf{$n=5$} \\
    \midrule
    sigmoid & 88.44  & 86.73 & 88.13  & 88.51  \\
    tanh  & 81.48  &  80.44 & 81.48  & 81.59  \\
    elu ($\alpha=1$)&   91.38 &  90.93 &  90.95 & 90.91  \\

    \bottomrule
    \end{tabular}%
  \label{tab1}%
\end{table}%

\begin{figure}[!htpb]
\centering
\includegraphics[scale = 0.25]{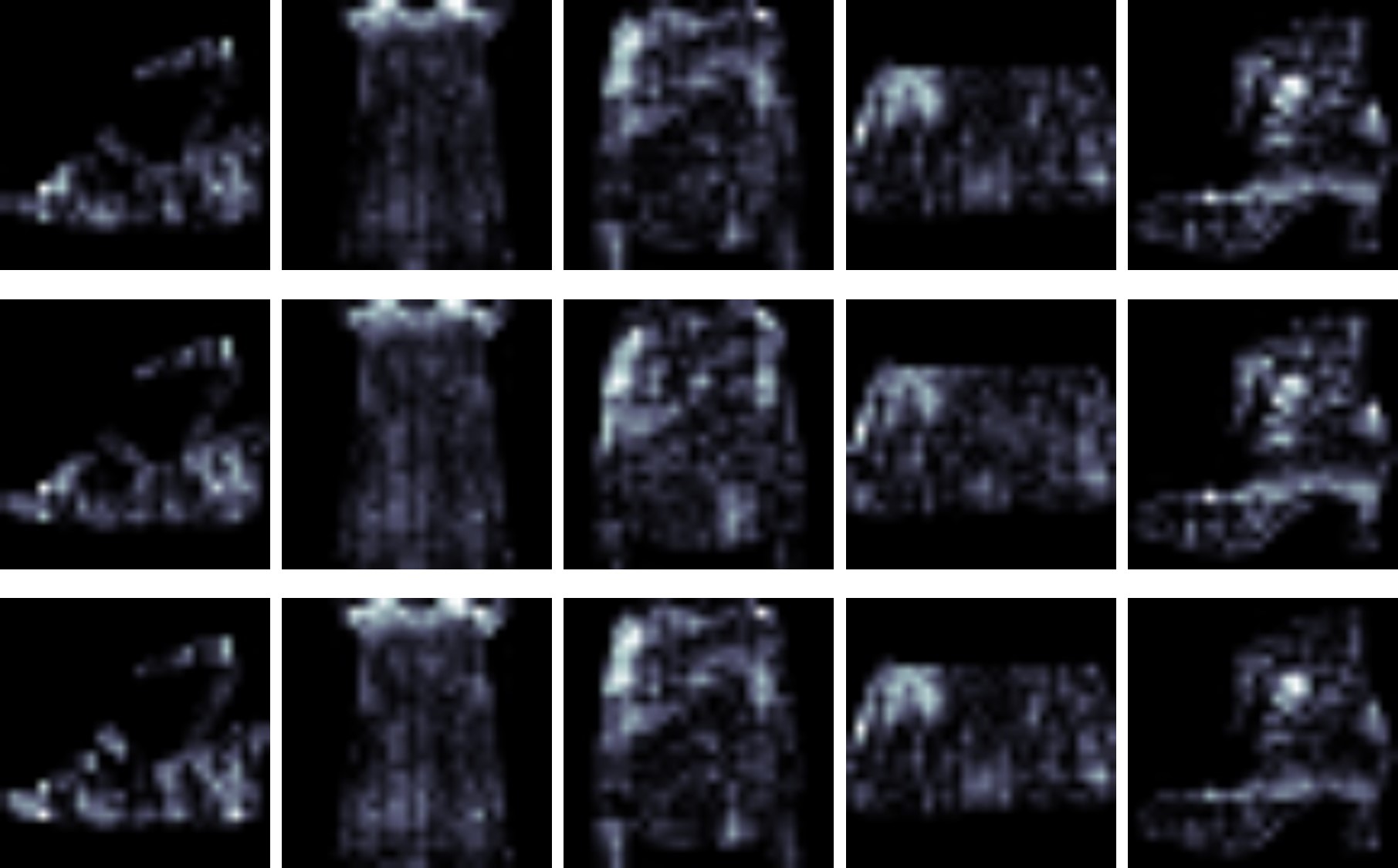}
\caption{ The comparison of the saliency maps~\cite{saliency1} (input gradient$\times$ input) between FCNN and the approximation models. \textbf{First Row}: sigmoid, \textbf{Second Row}: $n=3$, \textbf{Third Row}: $n=5$ ($n$ is the number of segments).}
\label{fig6}
\end{figure}

The saliency maps of the original model and the approximation ones are shown in Fig.\ref{fig6}. We can see that the approximation network has a semantic meaning that is extremely similar to the origin, supporting the efficacy of our method.

%\begin{figure}[!htb]
%\centering
%\subfloat[sigmoid]{
%\label{Fig.sub.1}
%\includegraphics[scale = 0.25]{fig6-sigmoid.jpg}}
%\subfloat[n=3]{
%\label{Fig.sub.2}
%\includegraphics[scale = 0.25]{fig6-3.jpg}}
%\subfloat[n=5]{
%\label{Fig.sub.3}
%\includegraphics[scale = 0.25]{fig6-5.jpg}}
%\caption{(a) and (b) are the decision boundary of the FCNN with sigmoid and tanh functions %respectively, (b) and (d) show the decision boundary of the FCNN with $p_3$ linear %approximate function}
%\label{fig6}
%\end{figure}

\subsection{DNet}

\begin{figure}[!htb]
\centering
\includegraphics[scale = 0.45]{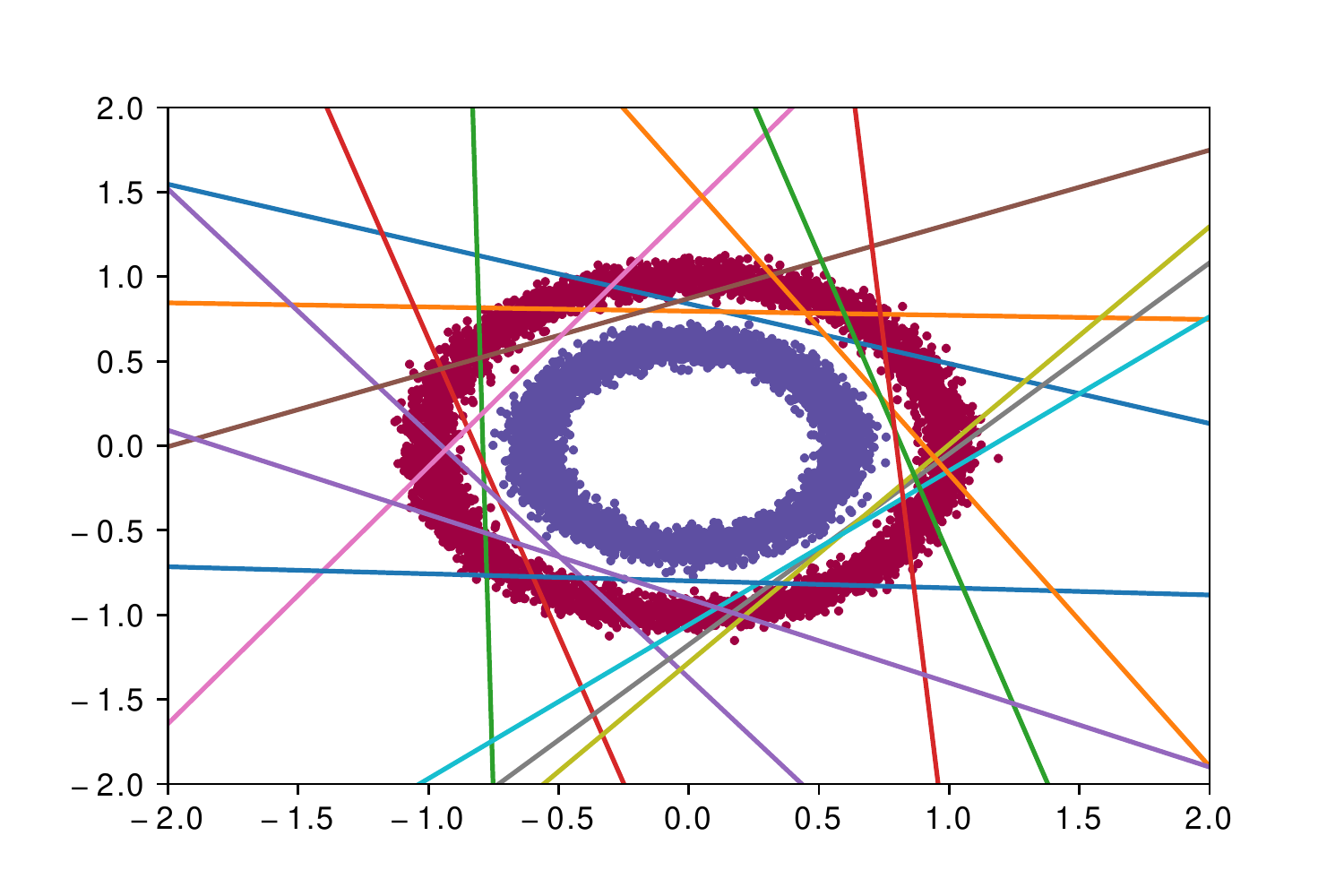}
\caption{ DNet on a toy dataset.}
\label{fig7}
\end{figure}

For intuition, we first explored the notion of DNet on a toy dataset~\cite{scikit-learn} with 10000 samples. DNet has the same accuracy (100\%) as the origin MLP (2-10-10-10-2) with ReLU activation, as demonstrated in Fig.\ref{fig7}. As a comparison, there are only 15 neurons required by DNet.

Furthermore, we created a sub-dataset of FashionMNIST with only two classes (T-shirt and Trouser). After that, we trained an MLP (784-10-10-4-2) with ReLU activation using a learning rate of 0.003 and a batch size of 128. The network converges after 10 epochs of training, at which point we compute its equivalent DNet. Finally, DNet (784-5) just requires 5 neurons to match the accuracy (99.80\%) of the original.

The effectiveness of DNet as demonstrated by the above experiments. We must emphasize drawbacks of DNet, which include the fact that it is not a traditional neural network and is hence less scalable.

\subsection{Distillation-based CLIME}

Convolution network is widely used in computer vision tasks, whose interpretability is more concerned. Therefore, we evaluate the performance of the distillation-based CLIME by interpreting the LeNet (The output layer is replaced with the softmax classifier). We use an MLP (784-256-64-10) with ReLU activation as student model, a well-trained LeNet as teacher model. We determine the best hypermeters by a grid search, where learning rate and batch size are form candidate sets [1e-5, 3e-5, 5e-5, 8e-5, 1e-4, 3e-4, 5e-4, 8e-4, 1e-3, 3e-3, 5e-3, 8e-3, 1e-2] and [32 ,64, 128, 256] respectively. Similarly, the hyperparameters of Hinton distillation are derived from two candidate sets: $\alpha$ from [0.5, 0.9, 0.95] and temperature $t$ from [1, 3, 5, 10, 20]. The results on two datasets are shown in Table~\ref{tab2}.

\begin{table}[htbp]
  \centering
  \caption{Test accuracy on MNIST and FashionMNIST datasets.}
    \begin{tabular}{lll}
    \toprule
    \textbf{Dataset} & \textbf{MNIST} & \textbf{FMNIST} \\
    \midrule
    LeNet & 98.93 & 89.16 \\
    \midrule
    MLP-baseline & 98.05 & 89.02 \\
    MLP-KD & \textbf{98.71} & \textbf{89.80 } \\
    \bottomrule
    \end{tabular}%
  \label{tab2}%
\end{table}%

\begin{figure}[!htpb]
\centering
\includegraphics[scale=0.6]{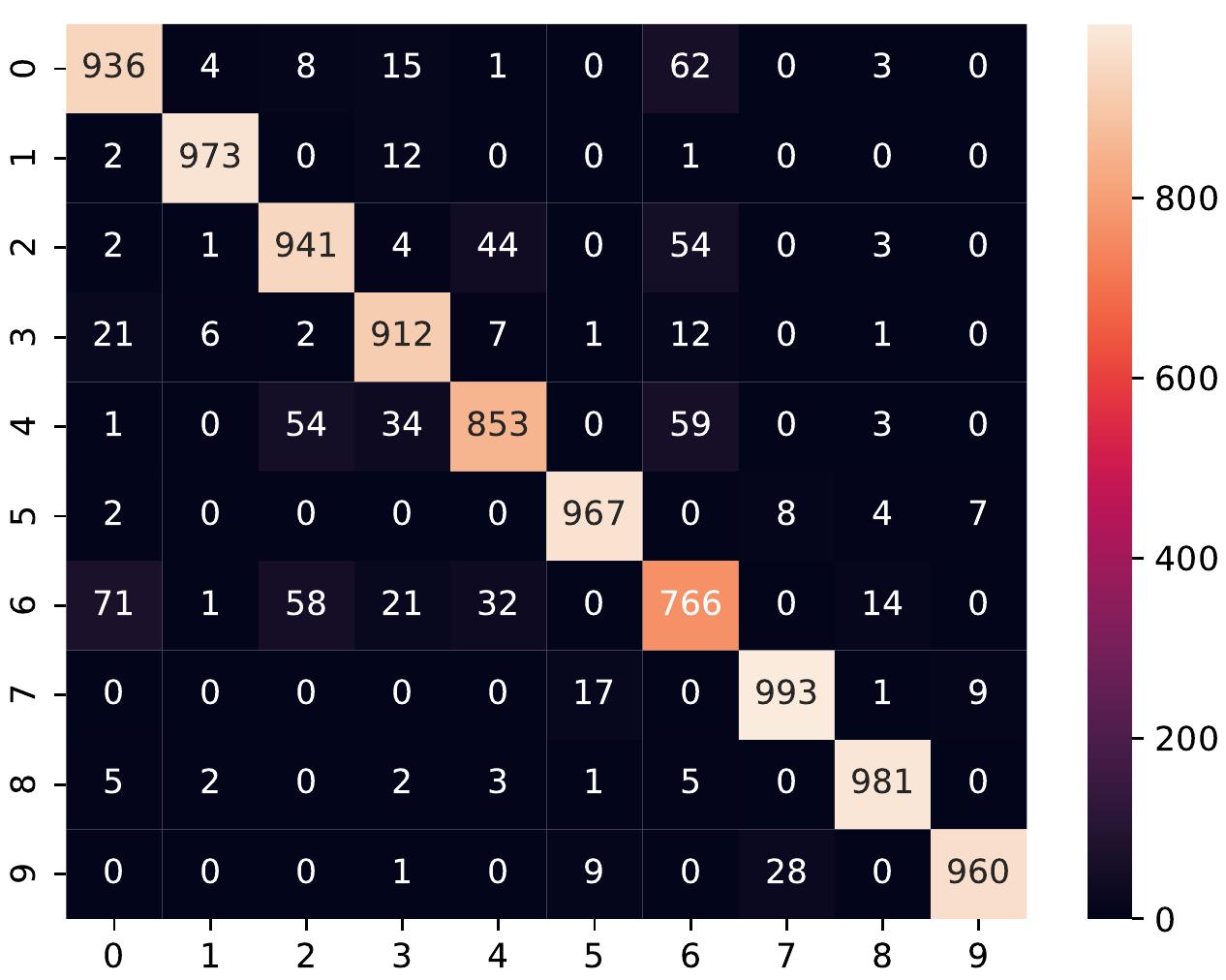} 
\caption{The confusion matrix between LeNet and MLP on FashionMNIST dataset. The horizontal axis is the LeNet classes, and the vertical one is the MLP classes.}
\label{fig8}
\end{figure}

The performance of MLP can be significantly boosted by distillation. Even on the FashionMNIST dataset, MLP dramatically beats LeNet. Of course, it is critical that the performance of LeNet does not significantly outperform MLP. Moreover, we plot the confusion matrix between LeNet and the MLP interpreter on the test set of FashionMNIST, as shown in Fig.\ref{fig8}. The MLP interpreter has 92.82\% of the same sample predictions as between LeNet, which confirms the validity of our distillation based CLIME.

Finally, we compare our distillation-based CLIME with LIME. Since LIME is difficult to handle high-dimensional data, we use LIME based on super-pixel. We randomly select 2000 samples in the test set of FashionMNIST, interpret and re-predict them using LIME, and then calculate the consistency with the output of LeNet. Consistency is measured by the proportion of samples with the same label to the total samples. As shown in Table.\ref{tab3}, our distillation-based CLIME substantially outperforms LIME, which demonstrates the advantages of using a more efficient interpreter.

% Table generated by Excel2LaTeX from sheet 'Sheet1'
\begin{table}[htbp]
  \centering
  \caption{Comparison of the consistency (\%) on FashionMNIST dataset. }
    \begin{tabular}{lrr}
    \toprule
    method & \multicolumn{1}{l}{LIME} & \multicolumn{1}{l}{CLIME} \\
    \midrule
    consistency & 9.75   & \textbf{93.45 } \\
    \bottomrule
    \end{tabular}%
  \label{tab3}%
\end{table}%

\section{Conclusions}
\label{sec:Conclusions}

We present a method to interpretation based on approximation theory in this study, which elucidates the fundamental issues in interpretability. We find piecewise linear approximations for fully-connected neural networks to construct an explanation. We propose the MLP interpreter to generalize this notion to arbitrary black-box models. The distillation-based interpreter, on the other hand, falls short of our need for complete approximation. Our trials also reveal that the correctness of the interpretation is more dependent on the performance of white-box model, indicating a direction for further study.

%%%%%%%%% REFERENCES

{\small
\bibliographystyle{unsrt}
\bibliography{egbib}

}

\end{document}